\documentclass[10pt,twocolumn,letterpaper]{article}

\usepackage{wacv}
\usepackage{times}
\usepackage{epsfig}

\usepackage{booktabs}
\usepackage{times}
\usepackage{soul}
\usepackage{url}
\usepackage[hidelinks]{hyperref}
\usepackage[utf8]{inputenc}
\usepackage[small]{caption}
\usepackage[dvipsnames]{xcolor}
\usepackage{multirow}
\usepackage{graphicx}
\usepackage{amsmath}
\usepackage{amssymb}
\usepackage{bm}

\wacvfinalcopy 

\def\wacvPaperID{760} 

\ifwacvfinal\fi
\begin{document}

\title{Animating Face using Disentangled Audio Representations}

\author{Gaurav Mittal \hspace{2cm} Baoyuan Wang \\
Microsoft Research\\
{\tt\small \{gamit, baoyuanw\}@microsoft.com 
}
}

\maketitle
\ifwacvfinal\thispagestyle{empty}\fi

\begin{abstract}
All previous methods for audio-driven talking head generation assume the input audio to be clean with a neutral tone. As we show empirically, one can easily break these systems by simply adding certain background noise to the utterance or changing its emotional tone (to such as sad). To make talking head generation robust to such variations, we propose an explicit audio representation learning framework that disentangles audio sequences into various factors such as phonetic content, emotional tone, background noise and others. We conduct experiments to validate that conditioned on disentangled content representation, the generated mouth movement by our model is significantly more accurate than previous approaches (without disentangled learning) in the presence of noise and emotional variations. We further demonstrate that our framework is compatible with current state-of-the-art approaches by replacing their original audio learning component with ours. To our best knowledge, this is the first work which improves the performance of talking head generation from disentangled audio representation perspective, which is important for many real-world applications. \footnote{Accepted at WACV 2020 (Winter Conference on Applications of Computer Vision)}
\end{abstract}

\section{Introduction}

With recent advances in deep learning, we have witnessed growing interest in automatically animating faces based on audio~(speech) sequences, thanks to applications in gaming, multi-lingual dubbing, virtual 3D avatars and so on. Specifically, the talking head generation is formulated as: given an input face image and an audio~(speech) sequence, the system needs to output a video where the mouth/lip region movement should be in synchronization with the phonetic content of the utterance while still preserving the original identity. 

As we all know, speech is riddled with variations. Different people utter the same word in different contexts with varying duration, amplitude, tone and so on. In addition to linguistic (phonetic) content, speech carries abundant information revealing about the speaker's emotional state, identity~(gender, age, ethnicity) and personality to name a few. Moreover, unconstrained speech recordings (such as from smartphones) inevitably contain certain amount of background noise.

There already exists a large body of research in the domain of talking head generation\cite{vougioukas2018end,zhou2019talking}. However, inspired by the rapid progress in Generative Adversarial Networks (GAN) \cite{goodfellow2014generative}, most of the recent works focus more on coming up with a better visual generative model to synthesize higher quality video frames. While impressive progress has been made by these prior methods, learning better audio representation specially tailored for talking head generation is being almost ignored without attracting much attention. For example, most of the previous works simply assume the input to be a clean audio sequence without any background noise or strong emotional tone, which is not unlikely in practical scenarios as we described above. We highlight in our empirical analysis that the state-of-the-art approaches are clearly unable to generalize to noisy and emotionally rich audio samples. 
Although recent works, such as \cite{Cocktail_Google} and \cite{audio-visual_enhancment}, show that visual signals can substantially help improve the audio quality (i.e remove noises) when the system can see the visual mouth movements, it is however reasonable to assume that in many cases video is not available, not to mention the misalignment issue between audio and video in practical online applications. 
 
Therefore, to make the system less sensitive to the noise, emotional tone and other potential factors, it is much desired to explicitly disentangle the audio representations first, before feeding into the talking head generation part, rather than simply treat it as black box and expect the network to implicitly handle the factors of variations. We argue that methods which explicitly decouple the various factors should have better chances to scale up the training and generalize well to unseen audio sequences, while implicit methods such as \cite{zhou2019talking,karras2017audio,Liu:2015,vougioukas2018end} may have high risk of over fitting.

To this end, we present a novel learning based approach to disentangle the phonetic content, emotional tone and other factors into different representations solely from the input audio sequence using the Variational Autoencoder\cite{kingma2014auto} framework. We encourage the decoupling by adding (1) local segment-level discriminative losses to regularize phonetic content representation, (2) global sequence-level discriminative loss to regularize the emotional tone and (3) margin ranking loss to separate out content from rest of the factors, in addition to the regular VAE losses. We further propose our own talking head generation module conditioned on the learned audio representation, in order to better evaluate the performance. To summarize, there are two major contributions of this work:
\begin{itemize}
    \item We present a novel disentangled audio representation learning framework for the task of generating talking heads. To the best of our knowledge, this is the first approach of improving the performance from audio representation learning perspective.
    \item Through various experiments, we show that our approach is not only robust to several naturally-existing audio variations but it is also compatible to be trained end-to-end with any of the existing talking head approaches.
\end{itemize}

\section{Related Work}
Speech-based facial animation literature can be broadly divided into two main categories. The first kind uses a blend of deep learning and computer graphics to animate a 3D face model based on audio.  \cite{Liu:2015} uses a data-driven regressor with an improved DNN acoustic model to accurately predict mouth shapes from audio. \cite{karras2017audio} performs speech-driven 3D facial animation mapping the input waveforms to 3D vertex coordinates of a face model and simultaneously using an emotional state representation to disambiguate the variations in facial pose for a given audio. \cite{Yang:2018:VisemeNet} introduces a deep learning based approach to map the audio features directly to the parameters of the JALI model \cite{Edwards:2016:JALI}. \cite{taylor2017deep} uses a sliding window approach to animate a parametric face model from phoneme labels. Recently, \cite{VOCA2019} introduced a model called Voice Operated Character Animation~(VOCA) which takes as input a speech segment in the form of its corresponding DeepSpeech~\cite{hannun2014deep} features and a one-hot encoding over training subjects to produce offsets for 3D face mesh for subject template registered using FLAME~\cite{FLAME:SiggraphAsia2017} model. Their approach is for 3D facial animation which allows altering speaking style, pose and shape, but cannot adapt completely to an unseen identity. The paper suggests DeepSpeech features to be robust to noise but we later show that these are not as efficient as our disentangled representations which are modeled to decouple from content not just noise but also other variations including emotion and speaking style.

The second category includes approaches performing audio-based 2D facial video synthesis, commonly called ``talking head/face generation". \cite{Chung17b} learns a joint audio-visual embedding using encoder-decoder CNN model and \cite{fan2015photo} uses Bi-LSTM to generate talking face frames.~\cite{suwajanakorn2017synthesizing} and \cite{kumarobamanet} both generating talking head for specifically Barack Obama using RNN with compositing techniques and time-delayed LSTM with pix2pix~\cite{isola2017image} respectively. \cite{jalalifar2018speech} uses RNN with conditional GAN and \cite{vougioukas2018end} uses Temporal GAN to synthesize talking faces. \cite{chen2018lip} employs optic-flow information between frames to improve photo-realism in talking heads.~\cite{zhou2019talking} proposes arbitrary-subject talking face generation using disentangled audio-visual representation with GANs.

Almost all of the previous approaches have been trained to work on clean neutral audio and fail to take into account many of the factors of variations occurring in real-world speech such as noise and emotion. Several recent works {} have demonstrated the importance of disentangled and factorized representation to learn a more generalized model~\cite{hsu2017unsupervised}. To the best of our knowledge, our approach is the first attempt to explicitly learn emotionally and content aware disentangled audio representations for facial animation. Some previous approaches \cite{karras2017audio,zhou2019talking,Liu:2015} do try to perform some kind of disentanglement but none of them explicitly deals with disentangling the different factors of variation in audio.  

\begin{figure*}[h]
    \centering
    \includegraphics[width=\textwidth]{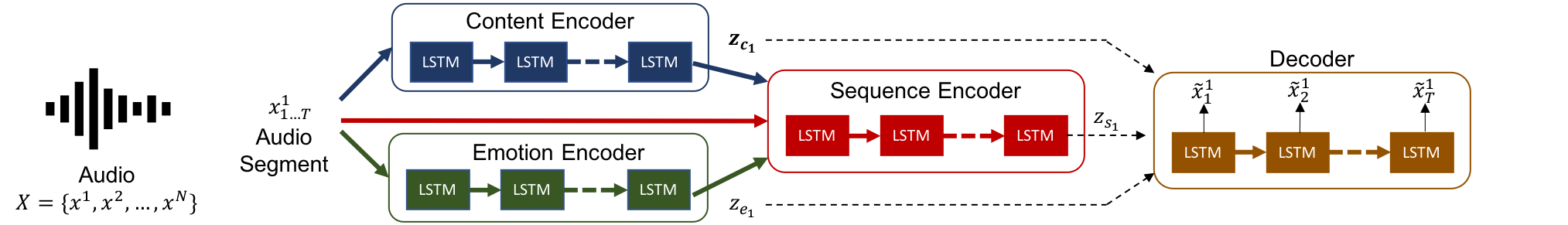}
    \caption{VAE architecture to learn emotionally and content aware disentangled audio representations}
    \label{fig:vae_archi}
\end{figure*}

\begin{figure}[h]
    \centering
    \includegraphics[width=\columnwidth]{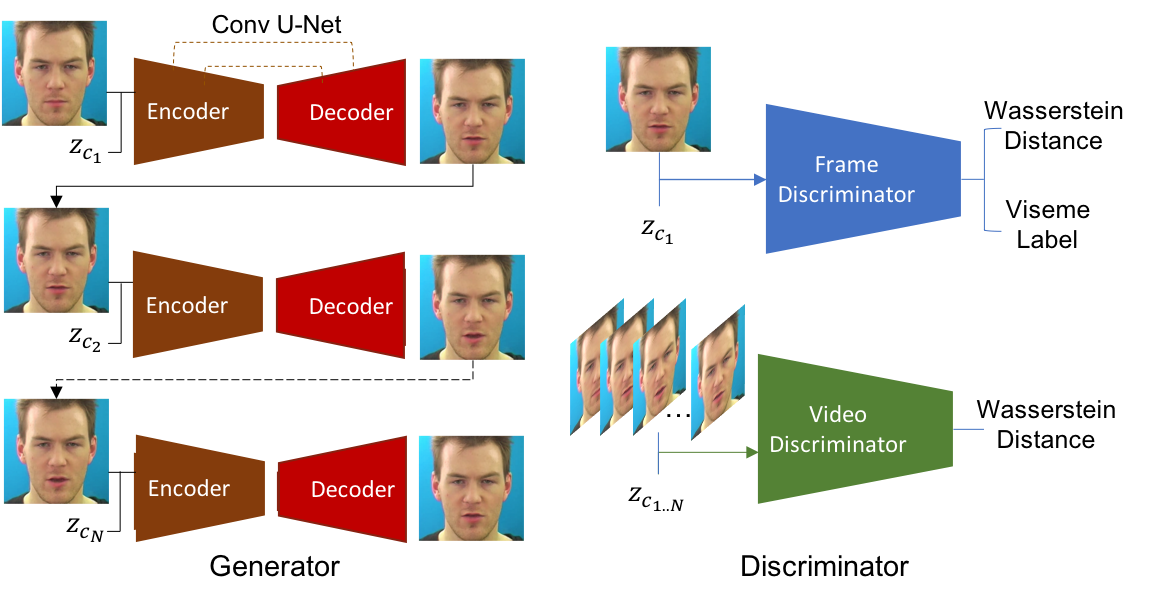}
    \caption{GAN based talking head generation model}
    \label{fig:gan_archi}
\end{figure}

\section{Method}
Our proposed method consists of two main stages,

\paragraph{Learning Disentangled Representations from Audio} The input audio sequence is factorized by a VAE into different representations encoding content, emotion and other factors of variations (Figure~\ref{fig:vae_archi}). KL divergence, negative log likelihood along with margin ranking losses ensure the learned representations are indeed disentangled and meaningful.

\paragraph{Generating Talking Head} Based on the input audio, a sequence of content representations are sampled from the learned distribution which along with the input face image are fed to a GAN-based video generator to animate the face (Figure~\ref{fig:gan_archi}). We use temporal smoothing along with frame and video discriminator~\cite{chan2018everybody,vougioukas2018end} here but as we show later, our audio representations are compatible with any existing talking head approach.

\subsection{Learning Disentangled Representations}

Speech comprises of several factors which act independently and at different temporal scales. Taking inspiration from \cite{hsu2017unsupervised}, we intend to disentangle content and emotion in an interpretable and hierarchical manner. We introduce several talking head generation specific novelties which include \textit{lateral disentanglement} of content and emotion by explicit decoupling using margin ranking losses, and a mechanism to learn \textit{variation-specific} priors which, unlike \cite{hsu2017unsupervised}, may or may not be sequence agnostic.  Syllables (linguistic content of an utterance) last only for few hundred milliseconds and do not exhibit significant variation within and between different speech sequences. We call this short duration a \textit{segment} and encode syllables by a set of \textit{latent content variables} regularized by a content-specific (viseme-oriented) prior that is sequence-independent. Since emotion is similar within a subset of utterances, we model emotion-related factors with \textit{latent emotion variables} regularized by a prior shared among sequences with the same emotion annotation. Finally, we need \textit{latent sequence variables} to encode residual variations of an entire utterance (\textit{sequence}) that can’t be captured by either content or emotion based variables. 

\paragraph{Model Formulation}
Let $\mathcal{D} = \lbrace \mathbf{X}^i \rbrace^{M}_{i=1}$ consists of M i.i.d. sequences where every $\mathbf{X^i} = \lbrace \bm{x}^{i,n}\rbrace^{N^i}_{n=1}$ is a sequence of $N^i$ observed variables with $N^i$ referring to the number of content segments (syllables) in the $i^{th}$ sequence and $\bm{x}^{i,n}$ referring to the $n^{th}$ content segment in the $i^{th}$ sequence. We omit $i$ in subsequent notations to refer to terms associated with a single sequence without loss in generality. 

Let each audio sequence $\mathbf{X}$ be randomly generated from a content-specific prior $\bm{\mu_{c}}$, emotion-specific prior $\bm{\mu_{e}}$ and sequence-specific prior $\bm{\mu_{s}}$ with $N$ i.i.d latent variables for content $\mathbf{Z_{c}}$, emotion $\mathbf{Z_{e}}$ and sequence $\mathbf{Z_{s}}$ (one for each of the $N$ segments in $X$). The joint probability for a sequence is therefore given by,

{\small
\begin{gather}
    p_\theta(\mathbf{X}, \mathbf{Z_{c}}, \mathbf{Z_{e}}, \mathbf{Z_{s}}, \bm{\mu_{c}}, \bm{\mu_{e}}, \bm{\mu_{s}})  =  p_\theta(\bm{\mu_{c}}) p_\theta(\bm{\mu_{e}}) p_\theta(\bm{\mu_{s}}) \notag  \\ 
     \prod^N_{n=1} p_\theta(\bm{x}^n|\bm{z}_{c}^n, \bm{z}_{e}^n, \bm{z}_{s}^n)   
     p_\theta (\bm{z}_{c}^n | \bm{\mu_{c}}) p_\theta (\bm{z}_{e}^n | \bm{\mu_{e}}) p_\theta (\bm{z}_{s}^n | \bm{\mu_{s}})
\end{gather}
}

where the priors $\bm{\mu_{c}}$, $\bm{\mu_{e}}$ and $\bm{\mu_{s}}$ are drawn from prior distributions $p_\theta(\bm{\mu_{c}})$, $p_\theta(\bm{\mu_{e}})$ and $p_\theta(\bm{\mu_{s}})$ respectively and the latent variables $\bm{z}_{c}^n$, $\bm{z}_{e}^n$ and $\bm{z}_{s}^n$ are drawn from isotropic multivariate Gaussian centred at $\bm{\mu_{c}}$, $\bm{\mu_{e}}$ and $\bm{\mu_{s}}$ respectively. $\theta$ represents the parameters of the generative model and the conditional distribution of $\bm{x}$ (audio segment) is modeled as a multivariate Gaussian with a diagonal covariance matrix.

Since the exact posterior inference is intractable, we use Variational Autoencoder~(VAE) to approximate the true posterior $p_\theta$ with an inference model $q_\phi$ given by,

{\small
\begin{gather}
    q_\phi(\mathbf{Z_{c}}^i, \mathbf{Z_{e}}^i, \mathbf{Z_{s}}^i, \mathbf{\mu}_c^i, \mathbf{\mu}_e^i, \mathbf{\mu}_s^i | \mathbf{X}^i)  =   q_\phi(\bm{\mu}_e^i) q_\phi(\bm{\mu}_s^i)  \notag \\
     \prod^N_{n=1} q_\phi(\bm{\mu}_c^{i,n})  q_\phi(\bm{z}_{s}^{i,n} | \bm{z}_{c}^{i,n}, \bm{z}_{e}^{i,n}, \bm{x}^{i,n}) \notag \\
   q_\phi (\bm{z}_{c}^{i,n} | \bm{x}^{i,n}) q_\phi (\bm{z}_{e}^{i,n} | \bm{x}^{i,n})
\label{inference}
\end{gather}}

\cite{hsu2017unsupervised} suggests that the mean $\bm{\tilde{\mu}_s}^i$ (one for each sequence) of $q_\phi(\bm{\mu}_s^i)$ be part of a lookup table and learned like other model parameters. We extend this idea to talking head scenario by introducing lookup tables for $q_\phi(\bm{\mu}_c^{i,n})$ and $q_\phi(\bm{\mu}_e^i)$ having different values $\bm{\tilde{\mu}_c}^{i,n}$ and $\bm{\tilde{\mu}_e}^i$ for different viseme and emotion labels respectively. Being at sequence level, $\bm{\tilde{\mu}_s}^{i,n} = \bm{\tilde{\mu}_s}^i$ and $\bm{\tilde{\mu}_e}^{i,n} = \bm{\tilde{\mu}_e}^i$ $\forall$  $n$. Based on the annotation, the corresponding value is picked up from the respective tables for optimization. Such \textit{variation-specific} priors allow the latent variables to be modeled effectively with samples with similar viseme/emotion made to lie closer together on the latent manifold. By further aligning $\bm{z}_s$ with $\bm{\tilde{\mu}_s}^i$, we encourage $\bm{z}_s$ to encode sequence-specific attributes which have larger variance across sequences but little variance within sequences.

The variational lower bound for this inference model over the marginal likelihood of $\textbf{X}$ is given as,
{\small
\begin{gather}
    \log p_\theta(\bm{X}) \geq  \mathcal{L}(\theta, \phi ; \bm{X}) =  \sum^{N}_{n=1} \left[ \mathcal{L}(\theta, \phi;\bm{x}^n | \bm{\tilde{\mu}}_c^n, \bm{\tilde{\mu}}_e, \bm{\tilde{\mu}}_s) \right.\notag \\
    \left. +  \log p_\theta(\bm{\tilde{\mu}}_c^n) \right]  + \log p_\theta(\bm{\tilde{\mu}}_e)
      + \log p_\theta(\bm{\tilde{\mu}}_s) + const  
\label{elbo}
\end{gather}}
Please refer to the supplementary material for proofs and a more detailed explanation of this section.
\paragraph{Discriminative Objective}
It is possible for the priors to learn trivial values for all sequences ($\bm{\Tilde{\mu}_c}^i = 0$, $\bm{\Tilde{\mu}_e}^i = 0$, $\bm{\Tilde{\mu}_s}^i = 0$ $\forall i$) and still maximize the variational lower bound described above. To ensure that the priors are indeed discriminative and characteristic of the variations they encode, we introduce a discriminative objective function that infers \textit{variation-specific} annotation from the corresponding representation. For instance, we enforce the content latent variable  $\bm{z}_c^{i,n}$ for audio segment $\bm{x}^{i,n}$ to correctly infer its viseme annotation $v^{i,n}$ through classification loss given by, 

{\small
\begin{equation}
\log p(v^{i,n}|\bm{z}_c^{i,n}) = \log p(\bm{z}_c^{i,n}|v^{i,n}) - \log \sum^V_{j=1} p(\bm{z}_c^{i,n}|v^{i,j})
\end{equation}
}

where $V$ is the set of all viseme labels~\cite{edwards2016jali}. We similarly enforce discriminative objective over emotion and sequence latent variables to correctly predict the emotion and sequence id associated with the audio sequence.

\paragraph{Margin Ranking Loss} 
To enable effective mapping of the audio content with the facial features and minimize ambiguity, we need to separate out the content from the rest of the factors of variations as much as possible. So we need to ensure that $\bm{z}_c$, $\bm{z}_e$ and $\bm{z}_s$ are as decoupled as possible. The discriminative objectives over the different latent variables ensure that they capture well their respective factors of variations (content, emotion and global sequence variations respectively) but to really disentangle them, we want to make them agnostic to other variations by having them perform badly on other classification tasks (that is, content variable $\bm{z}_c$ perform poorly in predicting the correct emotion associated with the audio sequence). To this end, we introduce margin ranking losses $\mathcal{T}$ with margin $\gamma$ on the softmax probability scores of the viseme label for $\bm{z}_c$ with $\bm{z}_s$ and $\bm{z}_e$ given by,

{\small
\begin{gather}
    \mathcal{T}(v^{i,n},\bm{z}_c^{i,n}n, \bm{z}_e^{i,n},\bm{z}_s^{i,n})  =  \max\left(0, \gamma + \mathcal{P}(v^{i,n}| \bm{z}_s^{i,n}) - \right. \notag \\ \left. \mathcal{P}(v^{i,n}| \bm{z}_c^{i,n})\right)  
     + \max\left(0, \gamma + \mathcal{P}(v^{i,n}| \bm{z}_e^{i,n}) -  \mathcal{P}(v^{i,n}| \bm{z}_c^{i,n})\right)
\end{gather}
}
where $\mathcal{P}(v^{i,n}| .)$ denotes the probability of $v^{i,n}$ given some latent variable. Margin ranking loss widens the inference gap, effectively making only $\bm{z}_c$ learn the content relevant features. We similarly introduce margin ranking loss on probability scores for emotion label to allow only $\bm{z}_e$ learn emotion relevant features. 

Equation~\ref{elbo} suggests that the variational lower bound of an audio sequence can be decomposed into the sum of variational lower bound of constituent segments. This provides scalability by allowing the model to train over audio segments instead. As shown in Figure~\ref{fig:vae_archi}, the input to the VAE is audio segments each having $T$ time points. Based on the inference model in Equation~\ref{inference}, these segments are first processed by LSTM-based content and emotion encoders, and later by sequence encoder (along with other latent variables). All the latent variables are then fed to the decoder to reconstruct the input. The final segment based objective function to maximize is as follows,

{\small
\begin{gather}
    \mathcal{L}^{F}(\theta, \phi; \bm{x}^{i,n})  =  \mathcal{L}(\theta, \phi;\bm{x}^{i,n})  - \beta [\mathcal{T}(e^i,\bm{z}_c^{i,n}, \bm{z}_e^{i,n},\bm{z}_s^{i,n}) \notag \\ +  \mathcal{T}(v^{i,n},\bm{z}_c^{i,n}, \bm{z}_e^{i,n},\bm{z}_s^{i,n})] + 
     \alpha [\log p(i|\bm{z}_s^{i,n}) \notag \\ + \log p(v^{i,n}|\bm{z}_c^{i,n}) 
     + \log p(e^i|\bm{z}_e^{i,n}) ]   
\end{gather}
}

where $\alpha$ and $\beta$ are hyper-parameter weights. 

\subsection{Talking Head Generation}
We use adversarial training to produce temporally coherent frames animating a given face image conditioned on the content representation $\bm{z}_c$ as shown in Figure~\ref{fig:gan_archi}.

\subsection{Generator} 
Let $G$ denote the generator function which takes as input a face image $I_f$ and sequence of audio-based content representations $\lbrace \bm{z}_c^{n}\rbrace^{N}_{n=1}$ sampled from $\bm{Z}_c$ given an audio sequence $\bm{X} = \lbrace \bm{x}^{n}\rbrace^{N}_{n=1}$ having $N$ audio segments. $G$ generates a frame $O_f^n$ for each audio segment $\bm{x}^{n}$. 
Each $\bm{z}_c^{n}$ is combined with the input image by channel-wise concatenating the representation after broadcasting over the height and width of the image. The combined input is first encoded and then decoded by $G$ which has a U-Net~\cite{ronneberger2015u} based architecture to output a video frame with the face modified in correspondence to the speech content. 
For temporal coherency between consecutive generated video frames, we introduce temporal smoothing similar to ~\cite{chan2018everybody} by making $G$ generate frames in an auto-regressive manner. We employ L1 loss along with perceptual similarity loss and L2 landmark distance (mouth region) as regularization.

\subsection{Discriminator} 
We incorporate WGAN-GP~\cite{gulrajani2017improved} based discriminators which act as critic to evaluate the quality of the generated frames/videos. We introduce a frame-level discriminator $D_{frame}$ which computes the Wasserstein distance of each individual generated frame conditioned on the input content representation. The architecture of $D_{frame}$ resembles that of PatchGAN~\cite{isola2017image}. $D_{frame}$ is designed to behave as a multi-task critic network. It also evaluates the conditioning between the generated frame and content representation through an auxiliary classification network that predicts the correct viseme corresponding to the conditioned audio segment (content representation). The loss for this auxiliary network is given by cross-entropy loss over the set of viseme labels. 

We introduce a video-level discriminator $D_{video}$ similar to~\cite{vougioukas2018end} to enforce temporal coherency in the generated video. The architecture of $D_{video}$ is similar to $D_{frame}$ without the auxiliary viseme classification network and has a 3D convolutional architecture with time representing the third dimension. It takes as input a set of frames (real or generated) along with corresponding content representations (concatenated channel wise) and evaluates the Wasserstein distance estimate over the video distribution. By doing so, $D_{video}$ evaluates the difference in realism and temporal coherence between the distribution of generated sequences and real sequences. 

\begin{figure*}[h]
    \centering
    \includegraphics[width=\textwidth]{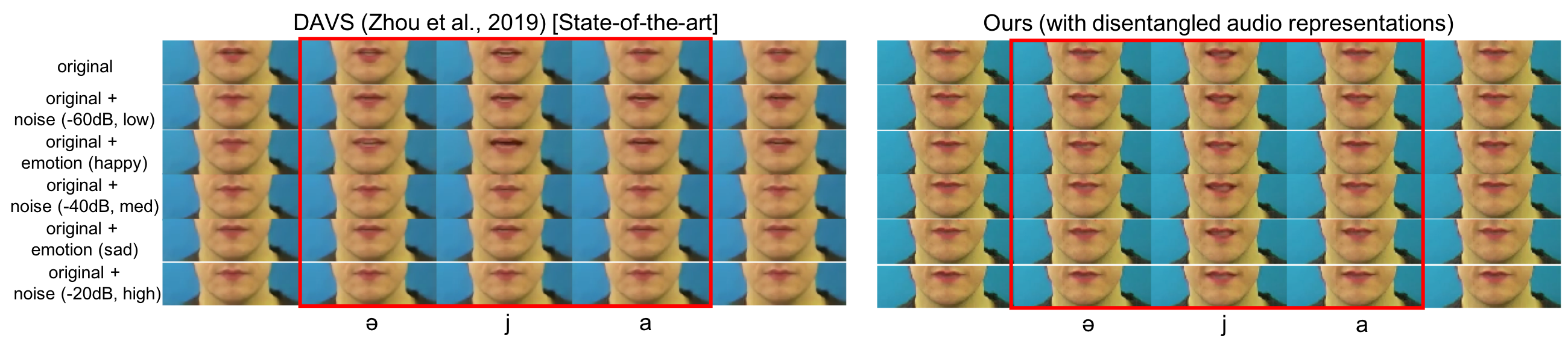}
    \caption{Visual comparison over different methods for different speech variations. If we look at the frames highlighted in the red box, we can observe how the introduction of noise or emotion reduces the performance/consistency of the current state-of-the-art while our approach is robust to such changes. Sentence: Don't forget a jacket. Symbols at the bottom denote syllables.}
    \label{fig:method_comparison}
\end{figure*}

\begin{figure*}[h]
    \centering
    \includegraphics[width=\textwidth]{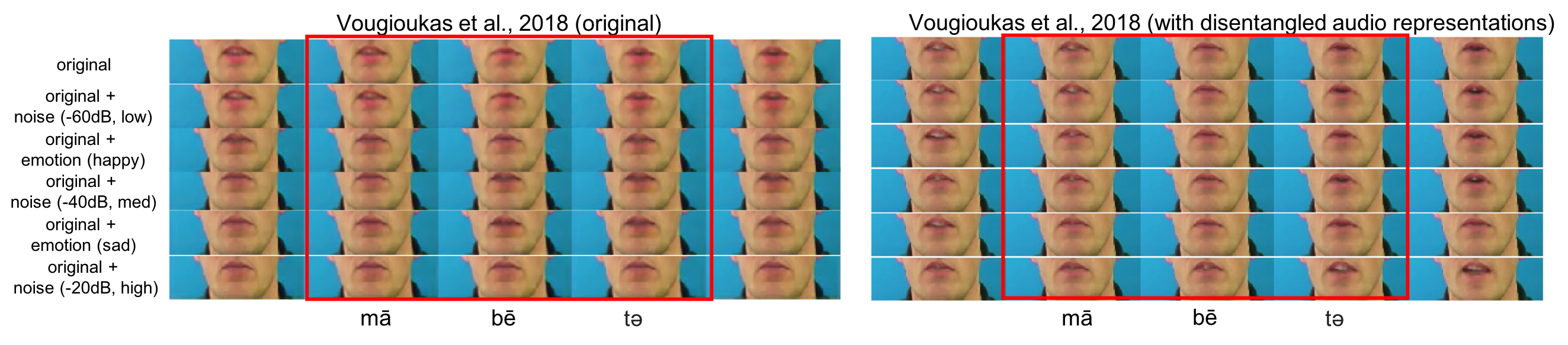}
    \caption{Visual comparison showing the ease of using our disentangled audio representation with existing talking head approaches to improve robustness to speech variations. Sentence: Maybe tomorrow it'll be cold. Symbols at the bottom denote syllables.}
    \label{fig:bmvc_compatibility}
\end{figure*}

\section{Experiments}

\subsection{Datasets}
\paragraph{GRID}\hspace{-5pt}\cite{cooke2006audio} is an audiovisual sentence corpus with high-quality recordings of 1000 sentences each from 34 talkers (18 male, 16 female) in a neutral tone. The dataset has high phonetic diversity but lacks any emotional diversity.
\paragraph{CRowdsourced Emotional Multimodal Actors Dataset (CREMA-D)}\hspace{-5pt}\cite{cao2014crema} consists of 7,442 clips from 91 ethnically-diverse actors (48 male, 43 female). Each speaker utters 12 sentences in 6 different emotions (Anger, Disgust, Fear, Happy, Neutral, Sad).
\paragraph{Lip Reading Sentence 3~(LRS3) Dataset}\hspace{-5pt}\cite{afouras2018lrs3} consists of over 100k spoken sentences from TED videos. We use this dataset to test our method in an `in-the-wild' audio-visual setting. Previous approaches have experimented with LFW~\cite{chung2016lip} which is a precursor to LRS3 dataset. 

\subsection{Training}
We use speech utterances from GRID and CREMA-D for training the VAE to learn disentangled representations. We divide the dataset speaker-wise using train-val-test split of 28-3-3 for GRID and 73-9-9 for CREMA-D. We first pre-train the content pipeline of the VAE using GRID (which provides the phonetic diversity) and then, use the learned weights to initialize the training of the entire VAE using CREMA-D (which provides the emotional diversity). To obtain the viseme annotations, we use Montreal Forced Aligner 
to extract phoneme annotation for each audio segment and then categorize them into 20 viseme groups (+1 for silence) based on~\cite{Yang:2018:VisemeNet}. Emotion labels are readily available from CREMA-D dataset for 6 different emotions. We label each audio sequence from GRID having neutral emotion.

We use a setup similar to~\cite{hsu2017unsupervised} for training the VAE. Every input speech sequence to the VAE is represented as a 200-dimensional log-magnitude spectogram computed every 10ms. Since the length of a syllabic segment is of the order of 200ms, we consider $\bm{x}$ to be a 200ms segment implying $T=20$ for each $\bm{x}$. We use 2-layer LSTM for all encoders and decoder with hidden size of $256$. Based on hyperparameter tuning, we set the dimensions for $\bm{z_c}, \bm{z_e}$ and $\bm{z_s}$ to 32, and the variance of priors to $1$ and latent variables to $0.25$. $\alpha$, $\beta$ and margin $\gamma$ are set to 10, 1 and 0.5 respectively. For generating talking head, we use GRID and LRS3 dataset. All faces in the videos are detected/aligned using~\cite{bulat2017far} and cropped to $256 \times 256$. Adam optimizer is used for training in both stages, and learning rate is fixed at $10^{-3}$ for VAE and $10^{-4}$ for GAN.

\begin{figure*}[h]
    \centering
    \includegraphics[width=\textwidth]{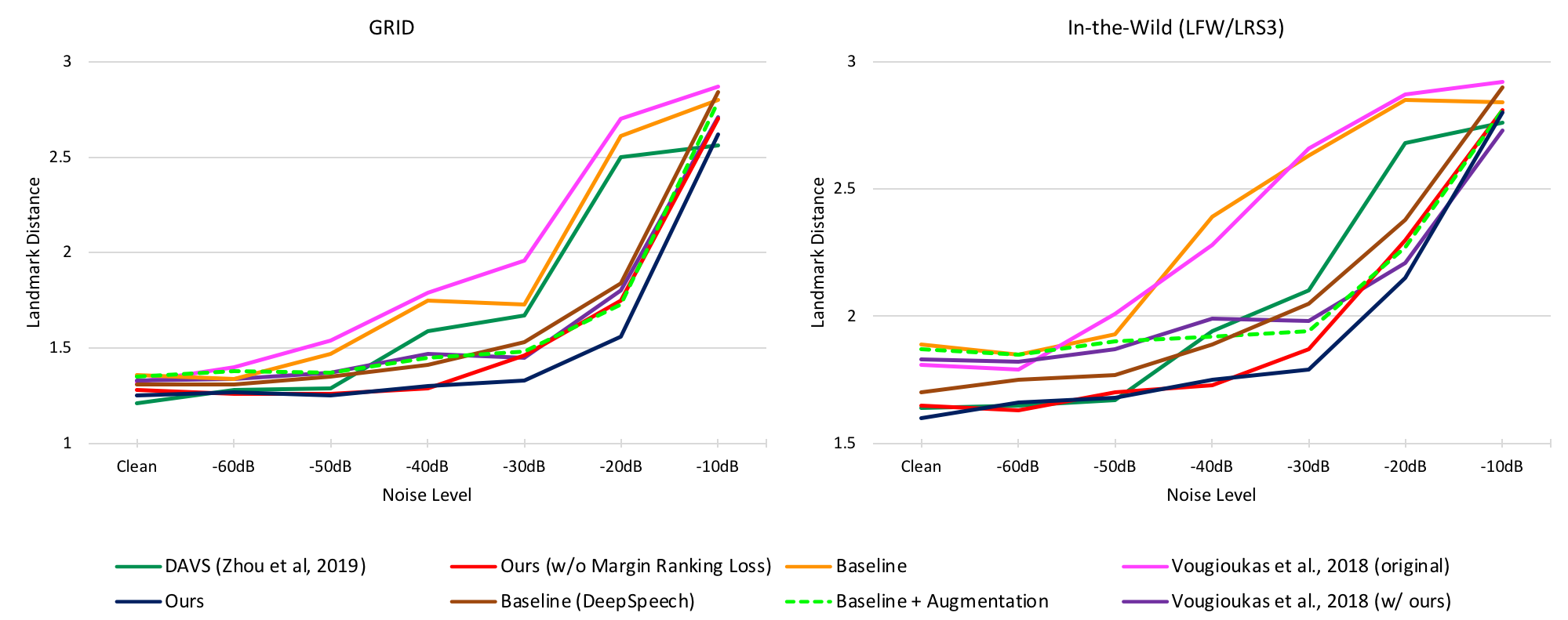}
    \caption{Plot for landmark distance comparison between different methods for different noise levels. Lower means better.}
    \label{fig:graph_noise}
\end{figure*}

\subsection{Robustness to Noise}
We evaluate the quality of the generated videos using Peak Signal to Noise Ratio (PSNR) and Structure Similarity Index Measure (SSIM). Higher the value of these metrics indicate better overall video quality. We further use Landmark Distance (LMD) (similar to \cite{chen2018lip}) to evaluate the accuracy of the mouth movement in the generated videos. LMD calculates the Euclidean Distance between the mouth landmarks as predicted by the landmark detection model \cite{bulat2017far} of the original video and the generated video.
{
\small
$$
LMD = \frac{1}{F} \times \frac{1}{L} \sum_{f=1}^{F} \sum_{l=1}^{L} ||P^{real}_{f,l} - P^{fake}_{f,l}||_2
$$
}

where $F$ denotes the number of frames in the video, $L$ denotes the number of mouth landmarks, and $P^{real}_{f,l}$ and $P^{fake}_{f,l}$ represents the landmark coordinates of the $l^{th}$ landmark in $f^{th}$ frame in the original and generated video respectively. Lower LMD denotes better talking head generation.

To test the robustness of our approach to noise, we create noisy samples by adding uniformly distributed white noise to audio sequences. We experiment with different noise levels by adjusting the loudness of the added noise compared to the original audio. A noise level of -40dB means that the added noise is 40 decibels lower in volume than the original audio. -10dB refers to high noise (almost imperceptible speech), -30dB refers to moderate (above average background noise) and -60dB refers to low noise (almost inaudible noise).

\begin{table}[h]
\footnotesize
\setlength\tabcolsep{0.8pt}
\begin{tabular}{lcccc|cccc}
\hline
\multicolumn{1}{c}{\multirow{2}{*}{Method}} & \multicolumn{4}{c}{GRID} & \multicolumn{4}{c}{LFW/LRS3} \\ \cline{2-9} 
\multicolumn{1}{c}{} & \multicolumn{1}{c}{Clean} & \multicolumn{1}{c}{\begin{tabular}[c]{@{}c@{}}Low\\ -60dB\end{tabular}} & \multicolumn{1}{c}{\begin{tabular}[c]{@{}c@{}}Med\\ -30dB\end{tabular}} & \multicolumn{1}{c}{\begin{tabular}[c]{@{}c@{}}High\\ -10dB\end{tabular}} & \multicolumn{1}{c}{Clean} & \multicolumn{1}{c}{\begin{tabular}[c]{@{}c@{}}Low\\ -60dB\end{tabular}} & \multicolumn{1}{c}{\begin{tabular}[c]{@{}c@{}}Med\\ -30dB\end{tabular}} & \multicolumn{1}{c}{\begin{tabular}[c]{@{}c@{}}High\\ -10dB\end{tabular}} \\ \hline
\cite{vougioukas2018end} (original) & 1.32 & 1.40 & 1.96 & 2.87 & 1.81 & 1.79 & 2.56 & 2.92 \\ 
\cite{vougioukas2018end} (w/ ours) & 1.33 & 1.34 & $\bm{1.45}$ & 2.71 & 1.83 & 1.82 & $\bm{1.98}$ & 2.73 \\ 
DAVS \cite{zhou2019talking} & 1.21 & 1.28 & 1.67 & 2.56 & 1.64 & 1.65 & 2.1 & 2.76 \\ 
Baseline & 1.36 & 1.34 & 1.73 & 2.80 & 1.89 & 1.85 & 2.63 & 2.84 \\ 
Baseline + Augmentation & 1.35 & 1.38 & 1.48 & 2.79 & 1.87 & 1.85 & 1.94 & 2.81 \\
Baseline (DeepSpeech) & 1.31 & 1.31 & 1.53 & 2.84 & 1.7 & 1.75 & 2.05 & 2.90 \\
Ours (w/o Margin Loss) & 1.28 & 1.26 & 1.46 & 2.7 & 1.65 & 1.63 & 1.87 & 2.81 \\ 
Ours & 1.25 & 1.27 & $\bm{1.33}$ & 2.62 & 1.67 & 1.66 & $\bm{1.79}$ & 2.80 \\ \bottomrule
\end{tabular}
\caption{Comparison of different approaches for audio samples with different noise levels.}
\label{noise_comparison}
\end{table}


Table~\ref{noise_comparison} shows the landmark distance estimates for different approaches over different noise levels. We re-implemented~\cite{vougioukas2018end} and used the public available model for DAVS~\cite{zhou2019talking} for obtaining and comparing the results. From the table, we can observe that for low noise levels, the performance of all the approaches is comparable to that for clean audio. But there is a significant rise in the landmark distance for \cite{vougioukas2018end} and DAVS as the noise levels become moderately high. While on the other hand, it is in this part of the noise spectrum where our approach excels and significantly outperforms the current state-of-the-art by maintaining a value comparable to clean audio. Clearly, by distentangling content from the rest of the factors of variations, our model is able to filter out most of the ambient noise and allow conditioning the video generation on a virtually cleaner signal. We observe that when the noise levels become exceedingly high, even our approach is unable to maintain its performance. We believe that such high noise levels completely distort the audio sequence leaving nothing meaningful to be captured and since we neither do any noise filtering nor use noisy samples for training explicitly, it is likely for the model to not perform well on almost imperceptible speech. Figure~\ref{fig:graph_noise} further shows a trend in the landmark distance for increasing noise levels. From the graph in Figure~\ref{fig:graph_noise}, we can observe that the performance of our approach becomes relatively better with increasing amounts of noise up to a reasonable level.

Figure~\ref{fig:method_comparison} shows a visual comparison of our approach with DAVS for different audio variations. We can notice for -40dB noise level, the mouth movement for DAVS begins to lose continuity with abrupt changes in the mouth movement (quick opening and closing of mouth) unlike for clean audio. By -20dB noise level, the mouth stops opening altogether. On the contrary, our method is much more resilient with mouth movement for -40dB noise level being almost identical to clean audio and for -20dB being only a bit abrupt.

\begin{table}[h]
\setlength\tabcolsep{4.8pt}
\begin{tabular}{llll|lll}
\hline
Method & \multicolumn{3}{c|}{GRID} & \multicolumn{3}{c}{LRW/LRS3} \\ \hline
       & LMD    & PSNR   & SSIM   & LMD          & PSNR         & SSIM         \\ \hline
\cite{vougioukas2018end}     &  1.32    & 28.88  & 0.81  &    1.81          &   28.49           & 0.71             \\ 
\cite{Chung17b}     &  1.35 & 29.36   & 0.74  &  2.25        & 28.06         &  0.46            \\ 
\cite{chen2018lip}   &   1.18 & 29.89   & 0.73  & 1.92         & 28.65         &  0.53           \\ 
\cite{zhou2019talking}   &   1.21      &   28.75     &    0.83   &    1.64           &  26.80       &   0.88         \\ 
Ours   &    1.25     &   30.43     &   0.78    &       1.67        &    29.12          &     0.73        \\ \bottomrule
\end{tabular}
\caption{Comparison with previous approaches on widely used metrics for original (clean) audio samples.}
\label{clean_comparison}
\end{table}

\begin{figure}[h]
    \centering
    \includegraphics[width=0.9\columnwidth]{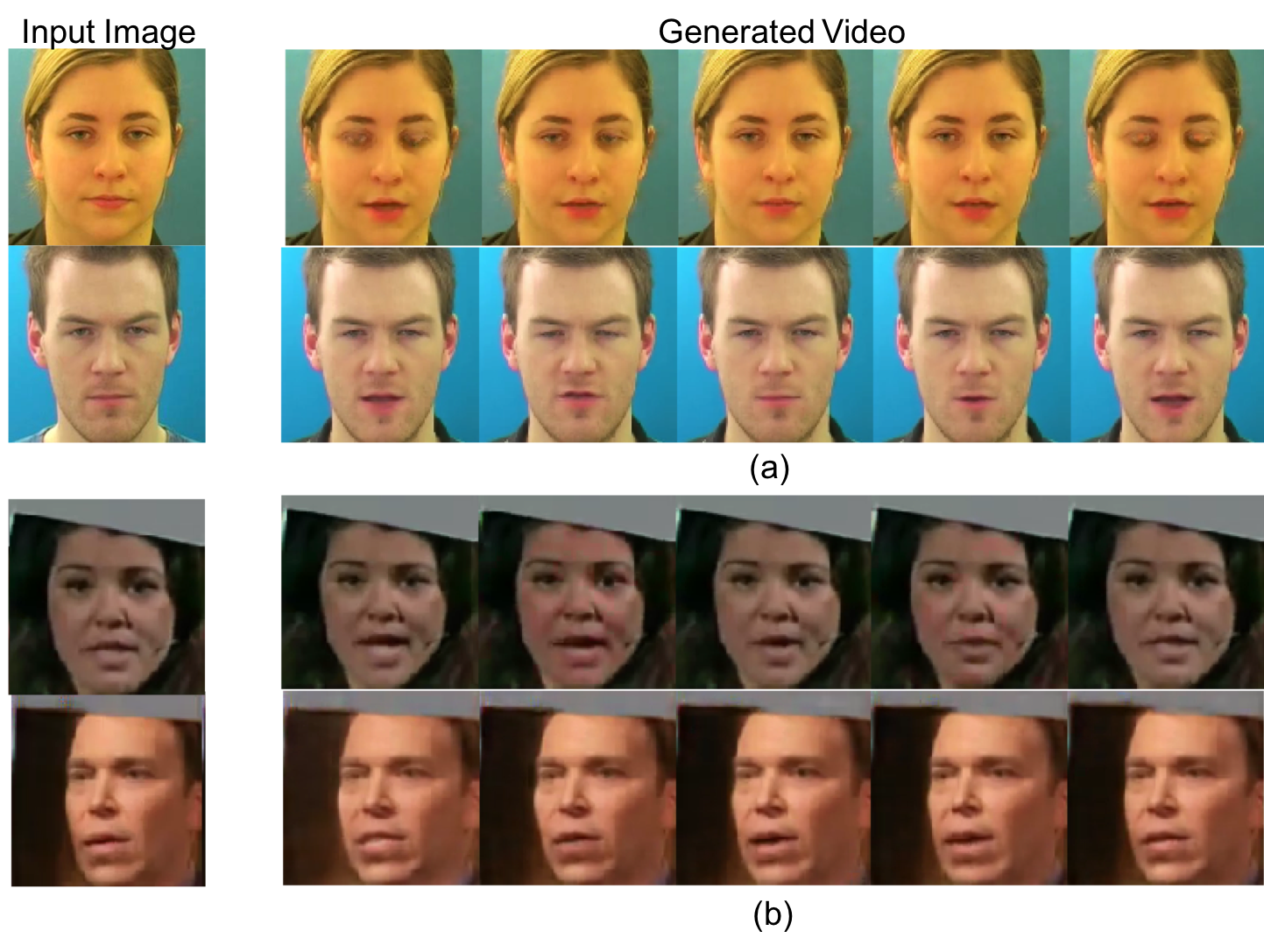}
    \vspace{-10pt}
    \caption{Sample results on (a) GRID and (b) LRS3 dataset for different speakers using clean audio samples.}
    \label{fig:more_results}
\end{figure}

We also show results of our approach on clean audio in Figure~\ref{fig:more_results}. Moreover from Table~\ref{clean_comparison}, we can observe that for clean neutral spoken utterances, our approach performs at par with other methods on all metrics.

\subsection{Robustness to Emotion}
We test the robustness of the disentangled representations to emotional variations by generating talking head for emotionally rich audio sequence from CREMA-D dataset. Due to this cross generation, we can only do a qualitative analysis as shown in Figure~\ref{fig:method_comparison}. We compare the talking head videos generated by our method with DAVS on different emotions. Looking at the frames in the red box, we can observe that although the performance of DAVS for loud emotions like happy is as good as for neutral, the mouth movement becomes abrupt and weak for soft emotions such as sad. On the contrary, our method is able to perform consistently over the entire emotional spectrum as evident from almost similar visual results for different emotions. 

\subsection{Ease of Compatibility}
Our model for learning emotionally and content aware disentangled audio representations is compatible with any of the current state-of-the-art approaches for talking head generation, and can be used in conjunction to improve robustness to factors of audio variations. We demonstrate this by implementing \cite{vougioukas2018end} using the content representation from our VAE model in place of that learned by the audio encoder. Table~\ref{noise_comparison} shows a comparison of the landmark distance between the two implementations for different noise levels. Similar to above, we can infer that using a filtered out content representation allows the model to perform significantly better than the original implementation in the presence of moderately high levels of noise.  From Figure~\ref{fig:graph_noise}, we can observe that the trend for the `hybrid' implementation is quite similar to our own implementation. Figure~\ref{fig:bmvc_compatibility} further compares the two implementations qualitatively for both noise and emotional audio. We can observe that \cite{vougioukas2018end} using our disentangled representations performs much more consistently than the original implementation. Due to the unavailability of training code/resources, we were unable to test our model with other approaches. But above demonstration proves that our disentangled representations can be easily incorporated with any existing implementation.

\subsection{Ablation study}
We conduct an ablation study to quantify the effect of each module in our approach. We run a baseline experiment where we replace our disentangled audio representation with a generic network which learns directly from MFCC features similar to~\cite{chung2016out}. As can be seen from Table~\ref{noise_comparison}, the baseline performs poorly for noisy samples. This clearly suggests that simple audio representation is not robust to audio variations while generating talking heads. 

We introduce a second baseline where we further perform aggressive augmentation of the input audio in the aforementioned baseline of learning directly from MFCC features. Figure~\ref{fig:graph_noise} and Table~\ref{triplet} show the results of these experiments (labeled Baseline + Augmentation). We observe that the landmark distance estimates are consistently better than the baseline without augmentation. However, these results are still noticeably worse than results of our approach. Data augmentation does make a difference over using normal dataset, however, we believe that simply relying on augmented data for training is not efficient enough as it is very challenging to augment the `right’ noise for the trained model to generalize well for real scenarios. 

To further test the effectiveness of the representation, we perform another baseline experiment where we replace the disentangled content features with speech features extracted from robust automatic speech recognition~(ASR) model, DeepSpeech~\cite{hannun2014deep}. Since \cite{VOCA2019} shows the noise robustness of DeepSpeech features while generating relative low-dimensional offsets of a 3D face mesh given an audio input, we wish to test their potential in generating in a visual space which is orders of magnitude higher in dimension. As shown in Figure~\ref{fig:graph_noise} and Table~\ref{triplet}, we find these speech features are not as effective as our disentangled audio representation for talking face generation. We believe the difference in performance is because the feature embedding from robust ASR models such as DeepSpeech is essentially a point embedding which because of being oriented towards solving a discriminative task loses a lot of key information about the variations in audio and can even be incorrect. Since we use a VAE, our content representation is modeled instead as a distribution which preserves these subtle variations by making it reconstruct the audio while aligning with audio content at the same time. This dual benefit (balance), which ASR models cannot offer, makes our content representation a much more informative and robust input for a high-dimensional generative task of face animation. 

\begin{table}[h]
\setlength\tabcolsep{5pt}
\begin{tabular}{ccc|cc}
\hline
\multirow{2}{*}{Representation} & \multicolumn{2}{c}{\begin{tabular}[c]{@{}c@{}}With Margin \\ Ranking Loss\end{tabular}} & \multicolumn{2}{c}{\begin{tabular}[c]{@{}c@{}}Without Margin \\ Ranking Loss\end{tabular}} \\ \cline{2-5}
 & Viseme & Emotion & Viseme & Emotion \\ \hline
Content & \textbf{77.1} & 24.5 & 58.7 & 37.0 \\ 
Emotion & 29.8 & \textbf{68.4} & 35.4 & 55.3 \\ \bottomrule
\end{tabular}
\caption{Accuracy (\%) over viseme and emotion classification task by disentangled content and emotion representations.}
\label{triplet}
\end{table}

In addition to learning a factorized audio representation, we also ensure an increased decoupling of the different representations by enforcing margin ranking loss as part of the training objective. Decoupling is essential to allow different audio variations to be captured exclusively by the designated latent variable which in turn helps in distilling the content information for improved robustness to variations.  To prove the importance of margin ranking loss, we evaluated the landmark distance metric of the model trained without margin ranking loss. From Figure~\ref{fig:graph_noise} and Table~\ref{triplet}, we can conclude that margin loss makes the approach robust to higher levels of noise, For GRID dataset, although for -40dB noise, the results for with/without margin ranking loss are comparable, there is a noticeable gap for -30dB noise level. Similar trend can also be observed for LRS-3/LFW dataset. We believe that although there is some level of disentanglement without margin ranking loss, when the audio is noisier, we need stronger disentanglement to produce more clear content representation which is possible due to margin ranking loss. To further quantify the effectiveness of margin ranking loss in decoupling, we train auxiliary classifiers over the content and emotion representations for the task of viseme and emotion classification. As shown in Table~\ref{triplet}, it is clearly evident that introduction of margin ranking loss makes the latent representation perform badly on tasks other than the designated task. In fact, it not only widens the performance gap between the representations for a particular task, but it also facilitates the designated representation to perform better than without margin ranking loss. 


\section{Conclusion and Future Work}
We introduce a novel approach of learning disentangled audio representations using VAE to make talking head generation robust to audio variations such as background noise and emotion. We validate our model by testing on noisy and emotional audio samples, and show that our approach significantly outperforms the current state-of-the-art in the presence of such audio variations. We further demonstrate that our framework is compatible with any of the existing talking head approaches by replacing the audio learning component in \cite{vougioukas2018end} with our module and showing that it is significantly robust than the original implementation. By adding margin ranking loss, we ensure that the factorized representations are indeed decoupled. Our approach to \textit{variation-specific} learnable priors is extensible to other speech factors such as identity and gender which can be explored as part of future work.

{\small
\bibliographystyle{ieee}
\bibliography{egbib}

\begin{thebibliography}{10}\itemsep=-1pt

\bibitem{audio-visual_enhancment}
T.~Afouras, J.~S. Chung, and A.~Zisserman.
\newblock The conversation: Deep audio-visual speech enhancement.
\newblock {\em CoRR}, 2018.

\bibitem{afouras2018lrs3}
T.~Afouras, J.~S. Chung, and A.~Zisserman.
\newblock Lrs3-ted: a large-scale dataset for visual speech recognition.
\newblock {\em arXiv preprint arXiv:1809.00496}, 2018.

\bibitem{bulat2017far}
A.~Bulat and G.~Tzimiropoulos.
\newblock How far are we from solving the 2d \& 3d face alignment problem? (and
  a dataset of 230,000 3d facial landmarks).
\newblock In {\em ICCV}, 2017.

\bibitem{cao2014crema}
H.~Cao, D.~G. Cooper, M.~K. Keutmann, R.~C. Gur, A.~Nenkova, and R.~Verma.
\newblock Crema-d: Crowd-sourced emotional multimodal actors dataset.
\newblock {\em IEEE transactions on affective computing}, 5(4):377--390, 2014.

\bibitem{chan2018everybody}
C.~Chan, S.~Ginosar, T.~Zhou, and A.~A. Efros.
\newblock Everybody dance now.
\newblock {\em arXiv preprint arXiv:1808.07371}, 2018.

\bibitem{chen2018lip}
L.~Chen, Z.~Li, R.~K~Maddox, Z.~Duan, and C.~Xu.
\newblock Lip movements generation at a glance.
\newblock In {\em Proceedings of the European Conference on Computer Vision
  (ECCV)}, pages 520--535, 2018.

\bibitem{Chung17b}
J.~S. Chung, A.~Jamaludin, and A.~Zisserman.
\newblock You said that?
\newblock In {\em BMVC}, 2017.

\bibitem{chung2016lip}
J.~S. Chung and A.~Zisserman.
\newblock Lip reading in the wild.
\newblock In {\em Asian Conference on Computer Vision}, pages 87--103.
  Springer, 2016.

\bibitem{chung2016out}
J.~S. Chung and A.~Zisserman.
\newblock Out of time: automated lip sync in the wild.
\newblock In {\em Asian conference on computer vision}, pages 251--263.
  Springer, 2016.

\bibitem{cooke2006audio}
M.~Cooke, J.~Barker, S.~Cunningham, and X.~Shao.
\newblock An audio-visual corpus for speech perception and automatic speech
  recognition.
\newblock {\em The Journal of the Acoustical Society of America}, 2006.

\bibitem{VOCA2019}
D.~Cudeiro, T.~Bolkart, C.~Laidlaw, A.~Ranjan, and M.~Black.
\newblock Capture, learning, and synthesis of {3D} speaking styles.
\newblock {\em Computer Vision and Pattern Recognition (CVPR)}, 2019.

\bibitem{Edwards:2016:JALI}
P.~Edwards, C.~Landreth, E.~Fiume, and K.~Singh.
\newblock Jali: an animator-centric viseme model for expressive lip
  synchronization.
\newblock {\em ACM Transactions on Graphics (TOG)}, 35(4):127, 2016.

\bibitem{edwards2016jali}
P.~Edwards, C.~Landreth, E.~Fiume, and K.~Singh.
\newblock Jali: an animator-centric viseme model for expressive lip
  synchronization.
\newblock {\em ACM Transactions on Graphics (TOG)}, 35(4):127, 2016.

\bibitem{Cocktail_Google}
A.~Ephrat, I.~Mosseri, O.~Lang, T.~Dekel, K.~Wilson, A.~Hassidim, W.~T.
  Freeman, and M.~Rubinstein.
\newblock Looking to listen at the cocktail party: {A} speaker-independent
  audio-visual model for speech separation.
\newblock {\em CoRR}, 2018.

\bibitem{fan2015photo}
B.~Fan, L.~Wang, F.~K. Soong, and L.~Xie.
\newblock Photo-real talking head with deep bidirectional lstm.
\newblock In {\em 2015 IEEE International Conference on Acoustics, Speech and
  Signal Processing (ICASSP)}, pages 4884--4888. IEEE, 2015.

\bibitem{goodfellow2014generative}
I.~Goodfellow, J.~Pouget-Abadie, M.~Mirza, B.~Xu, D.~Warde-Farley, S.~Ozair,
  A.~Courville, and Y.~Bengio.
\newblock Generative adversarial nets.
\newblock In {\em Advances in neural information processing systems}, pages
  2672--2680, 2014.

\bibitem{gulrajani2017improved}
I.~Gulrajani, F.~Ahmed, M.~Arjovsky, V.~Dumoulin, and A.~C. Courville.
\newblock Improved training of wasserstein gans.
\newblock In {\em Advances in Neural Information Processing Systems}, pages
  5767--5777, 2017.

\bibitem{hannun2014deep}
A.~Hannun, C.~Case, J.~Casper, B.~Catanzaro, G.~Diamos, E.~Elsen, R.~Prenger,
  S.~Satheesh, S.~Sengupta, A.~Coates, et~al.
\newblock Deep speech: Scaling up end-to-end speech recognition.
\newblock {\em arXiv preprint arXiv:1412.5567}, 2014.

\bibitem{hsu2017unsupervised}
W.-N. Hsu, Y.~Zhang, and J.~Glass.
\newblock Unsupervised learning of disentangled and interpretable
  representations from sequential data.
\newblock In {\em Advances in neural information processing systems}, 2017.

\bibitem{isola2017image}
P.~Isola, J.-Y. Zhu, T.~Zhou, and A.~A. Efros.
\newblock Image-to-image translation with conditional adversarial networks.
\newblock In {\em 2017 IEEE Conference on Computer Vision and Pattern
  Recognition (CVPR)}, pages 5967--5976. IEEE, 2017.

\bibitem{jalalifar2018speech}
S.~A. Jalalifar, H.~Hasani, and H.~Aghajan.
\newblock Speech-driven facial reenactment using conditional generative
  adversarial networks.
\newblock {\em arXiv preprint arXiv:1803.07461}, 2018.

\bibitem{karras2017audio}
T.~Karras, T.~Aila, S.~Laine, A.~Herva, and J.~Lehtinen.
\newblock Audio-driven facial animation by joint end-to-end learning of pose
  and emotion.
\newblock {\em ACM Transactions on Graphics (TOG)}, 36(4):94, 2017.

\bibitem{kingma2014auto}
D.~P. Kingma and M.~Welling.
\newblock Auto-encoding variational bayes.
\newblock {\em ICLR}, 2014.

\bibitem{kumarobamanet}
R.~Kumar, J.~Sotelo, K.~Kumar, A.~de~Br{\'e}bisson, and Y.~Bengio.
\newblock Obamanet: Photo-realistic lip-sync from text.

\bibitem{FLAME:SiggraphAsia2017}
T.~Li, T.~Bolkart, M.~J. Black, H.~Li, and J.~Romero.
\newblock Learning a model of facial shape and expression from {4D} scans.
\newblock {\em ACM Transactions on Graphics, (Proc. SIGGRAPH Asia)}, 36(6),
  2017.

\bibitem{Liu:2015}
Y.~Liu, F.~Xu, J.~Chai, X.~Tong, L.~Wang, and Q.~Huo.
\newblock Video-audio driven real-time facial animation.
\newblock {\em ACM Trans. Graph.}, 34(6):182:1--182:10, Oct. 2015.

\bibitem{ronneberger2015u}
O.~Ronneberger, P.~Fischer, and T.~Brox.
\newblock U-net: Convolutional networks for biomedical image segmentation.
\newblock In {\em International Conference on Medical image computing and
  computer-assisted intervention}, pages 234--241. Springer, 2015.

\bibitem{suwajanakorn2017synthesizing}
S.~Suwajanakorn, S.~M. Seitz, and I.~Kemelmacher-Shlizerman.
\newblock Synthesizing obama: learning lip sync from audio.
\newblock {\em ACM Transactions on Graphics (TOG)}, 36(4):95, 2017.

\bibitem{taylor2017deep}
S.~Taylor, T.~Kim, Y.~Yue, M.~Mahler, J.~Krahe, A.~G. Rodriguez, J.~Hodgins,
  and I.~Matthews.
\newblock A deep learning approach for generalized speech animation.
\newblock {\em ACM Transactions on Graphics (TOG)}, 36(4):93, 2017.

\bibitem{vougioukas2018end}
K.~Vougioukas, S.~Petridis, and M.~Pantic.
\newblock End-to-end speech-driven facial animation with temporal gans.
\newblock {\em arXiv preprint arXiv:1805.09313}, 2018.

\bibitem{zhou2019talking}
H.~Zhou, Y.~Liu, Z.~Liu, P.~Luo, and X.~Wang.
\newblock Talking face generation by adversarially disentangled audio-visual
  representation.
\newblock In {\em AAAI Conference on Artificial Intelligence (AAAI)}, 2019.

\bibitem{Yang:2018:VisemeNet}
Y.~Zhou, Z.~Xu, C.~Landreth, E.~Kalogerakis, S.~Maji, and K.~Singh.
\newblock Visemenet: Audio-driven animator-centric speech animation.
\newblock {\em ACM Transactions on Graphics}, 37(4), 2018.

\end{thebibliography}
}

\end{document}